\documentclass[a4paper]{article}

\usepackage[utf8]{inputenc}
\usepackage{hyperref}
\usepackage{ctable}
\usepackage{listings}
\usepackage[numbers]{natbib}
\usepackage[blocks]{authblk} 
\makeatletter

\renewcommand\bibsection%
{
\section*{\refname
\@mkboth{\MakeUppercase{\refname}}{\MakeUppercase{\refname}}}
}
\makeatother

\providecommand{\keywords}[1]{ \small\textbf{Keywords:~~} #1}

\usepackage{listings,xcolor}
\lstdefinelanguage{ttl}{
sensitive=true,
morekeywords={nif,itsrdf,es,dbo,penn,olia,rdfs},
morecomment=[l][\color{black}]{@},
morecomment=[l][\color{darkgreen}]{\# },
morecomment=[l][\color{black}]{\#char },
morestring=[b][\color{blue}]\",
}
\definecolor{darkgreen}{RGB}{0, 128, 0}
\definecolor{lightblue}{RGB}{50,155,230}
\definecolor{black}{RGB}{0,0,0}

\begin{document}

\title{DBpedia NIF: Open, Large-Scale and Multilingual Knowledge Extraction Corpus}

\author[1,2]{Milan Dojchinovski}
\author[1,3]{Julio Hernandez}
\author[1]{Markus Ackermann}
\author[1]{Amit Kirschenbaum}
\author[1]{Sebastian Hellmann}

\affil[1]{Agile Knowledge Engineering and Semantic Web (AKSW) \authorcr 
InfAI, Leipzig University, Germany\authorcr 
\texttt{\{dojchinovski,ackermann,amit,hellmann\}@informatik.uni-leipzig.de}}

\affil[2]{Web Intelligence Research Group\authorcr Faculty of Information Technology\authorcr Czech Technical University in Prague, Czech Republic\authorcr 
\texttt{milan.dojchinovski@fit.cvut.cz}}

\affil[3]{Center for Research and Advanced Studies\authorcr
National Polytechnic Institute of Mexico\authorcr
\texttt{nhernandez@tamps.cinvestav.mx}}

\date{}
\setcounter{footnote}{0}

\maketitle

\begin{abstract}
In the past decade, the DBpedia community has put significant amount of effort on developing technical infrastructure and methods for efficient extraction of structured information from Wikipedia.
These efforts have been primarily focused on harvesting, refinement and publishing semi-structured information found in Wikipedia articles, such as information from infoboxes, categorization information, images, wikilinks and citations.
Nevertheless, still vast amount of valuable information is contained in the unstructured Wikipedia article texts.
In this paper, we present \textit{DBpedia NIF} - a large-scale and multilingual knowledge extraction corpus.
The aim of the dataset is two-fold: to dramatically broaden and deepen the amount of structured information in DBpedia, and to provide large-scale and multilingual language resource for development of various NLP and IR task.
The dataset provides the content of all articles for 128 Wikipedia languages.
We describe the dataset creation process and the NLP Interchange Format (NIF) used to model the content, links and the structure the information of the Wikipedia articles.
The dataset has been further enriched with about 25\% more links and selected partitions published as Linked Data.
Finally, we describe the maintenance and sustainability plans, and selected use cases of the dataset from the TextExt knowledge extraction challenge.

\keywords{DBpedia, NLP, IE, Linked Data, training, corpus}
\end{abstract}

\section{Introduction}
In past decade, the Semantic Web community has put significant effort on developing technical infrastructure for efficient Linked Data publishing.
These efforts gave birth of 1,163 Linked Datasets (as of February 2017\footnote{\url{http://lod-cloud.net/}}), which is an overall growth of 296\% datasets published in September 2011.
Since the beginning of the Linked Open Data (LOD) initiative, DBpedia \cite{dbpedia} has served as a central interlinking hub for the emerging Web of Data.
The ultimate goal of the DBpedia project was, and still is, to extract structured information from Wikipedia and to make this information available on the Web.
In the past ten years, the main focus of the DBpedia project was to extract available structured information found in the Wikipedia articles, such as information from the infoboxes, categorization information, images, wikilinks and citation.
Nevertheless, huge amount of highly valuable information is still being hidden in the free text of the Wikipedia articles.

The Wikipedia article texts represent the largest part of the articles in terms of time spent on writing, informational content and size.
This content and the information it provides can be exploited in various use cases, such as fact extraction and validation, training various multilingual NLP tasks, development of multilingual language resources, etc.
In the past years, there have been several attempts \cite{richman2008mining,toral2006,kazama2007exploiting,mihalcea2007wikify,NOTHMAN2013151,Mendes2011spotlight, HAHM14,POLYGLOT} on extracting the free text content in Wikipedia articles.
However, none of these works have achieved significant impact and recognition within the Semantic Web and NLP community since the generated datasets have not been properly maintained and they have been developed without sustainable development and maintenance strategy.
Thus, there is need to develop robust extraction process for extraction of the information from the free text in the Wikipedia articles, semantically describe the extracted information and make it available to the Semantic Web community, and put in place sustainable development and maintenance strategy for the dataset.

In order to achieve these goals, we have developed the \textit{DBpedia NIF} dataset - a large-scale, open and multilingual knowledge extraction dataset which provides information extracted from the unstructured information found in the Wikipedia articles.
It provides the Wikipedia article texts of 128 Wikipedia editions.
The dataset captures the structure of the articles as they are organized in Wikipedia including the sections, paragraphs and corresponding titles.
We also capture the links present in the articles, the surface forms (anchors) and their exact location within the text.
Since the Wikipedia contributors should follow strict guidelines while adding new links, there are many missing links in the content.
In order to overcome the problem of missing links, we further enriched the content with new links and significantly increased the number links, for example, the number of links for the English Wikipedia has been increased by 31.36\%.
The dataset is provided in a machine-readable format, in the NLP Interchange Format (NIF) \cite{hellmann2013NIF}, which is used to semantically describe the structure of the text documents and the annotations .
The dataset has been enriched, selected partitions published according to the Linked Data principles and new updated versions of the datasets are provided along with each DBpedia release.
We have evaluated the data for its syntactic validity and semantic accuracy and the results from these evaluations confirm the quality of the dataset.

The reminder of the paper is structured as follows.
Section~\ref{sec.background-and-motivations} provides necessary background information and motivates the work.
Section~\ref{sec.dataset} describes the dataset, the extraction process, the data model, the availability, and the maintenance and sustainability plans for the dataset.
Section~\ref{sec.enrichment} describes the enrichment process of the dataset.
Section~\ref{sec.quality} discusses the quality of the dataset.
Section~\ref{sec.use-cases} presents selected use cases of the dataset at the TextExt challenge.
Finally,  Section~\ref{sec.conclusion} concludes the paper.

\section{Background and Motivations}
\label{sec.background-and-motivations}

DBpedia is a community project which aims at published structured knowledge extracted from Wikipedia.
Since its inception, the DBpedia project has been primarily focused on extraction of knowledge from semi-structured sections in Wikipedia articles, such as infoboxes, categorization information, images, wikilinks, etc.
Nevertheless, huge amount of information is still being hidden in the text of the Wikipedia articles.
This information can not only increase the coverage of DBpedia but it can also support other relevant tasks, such as validation of DBpedia facts or training various NLP and IR tasks.

In the past, Wikipedia and the article texts have been widely exploited as a resource for many NLP and IR tasks \cite{richman2008mining,toral2006,kazama2007exploiting,mihalcea2007wikify,NOTHMAN2013151,Mendes2011spotlight, HAHM14,POLYGLOT}.
In \cite{richman2008mining} the authors develop a multilingual NER system, which is trained on data extracted from Wikipedia article texts.
Wikipedia has been used to create an annotated training corpus in seven languages by exploiting the wikilinks within the text.
In \cite{toral2006} the authors propose use of Wikipedia for automatic creation and maintenance of  gazetteers for NER.
Similarly, \cite{kazama2007exploiting} used Wikipedia, particularly the first sentence of each article, to create lists of entities.
Another prominent work which uses Wikipedia is Wikify! \cite{mihalcea2007wikify}, a system for enrichment of content with Wikipedia links.
It uses Wikipedia to create a vocabulary with all surface forms collected from the Wikipedia articles.
DBpedia Spotlight \cite{Mendes2011spotlight} is a system for automatic annotation of text documents
with DBpedia URIs.
It is using Wikipedia article text to collect surface forms which are then used for the entity spotting and disambiguation tasks.
In \cite{HAHM14}, the authors also construct NE corpus using Wikipedia, parsed the XML dumps and extracted wikilinks annotations.
In \cite{NOTHMAN2013151}, the authors present an approach for automatic construction of an NER training corpus out of Wikipedia, which has been generated for nine languages based on the textual and structural features present in the Wikipedia articles.
Similarly, in \cite{POLYGLOT} the authors develop an NER system which is trained on data from Wikipedia.
The training corpus has been generated for 40 languages by exploiting the Wikipedia link stricture and the internal links embedded in the Wikipedia articles to detect named entity mentions.

Although these works have shown promising results and confirmed the high potential of the Wiki text, several crippling problems are starting to surface.
Most of the datasets are not available, they are not regularly updated nor properly maintained, and they have been developed without clear sustainability plan and roadmap strategy.
Most datasets are provided in just few languages, with exception of \cite{POLYGLOT} which is provided in 40 languages.
The data is not semantically described, it is hard to query, analyze and consume.
Thus, there is need to develop a knowledge extraction process for extraction of information from Wikipedia article texts, semantically describe the extracted information, make it available to the Semantic Web and other communities, and establish sustainable development and maintenance strategies for the dataset.

\section{The DBpedia NIF Dataset}
\label{sec.dataset}

The DBpedia knowledge base is created using the DBpedia Extraction Framework\footnote{\url{https://github.com/dbpedia/extraction-framework/}}.
In our work, we appropriately extended the framework and implemented a knowledge extraction process for extraction of information from Wikipedia article texts.

\subsection{Knowledge Extraction}

\begin{figure*}[t!]
\centering
\includegraphics[width=\textwidth]{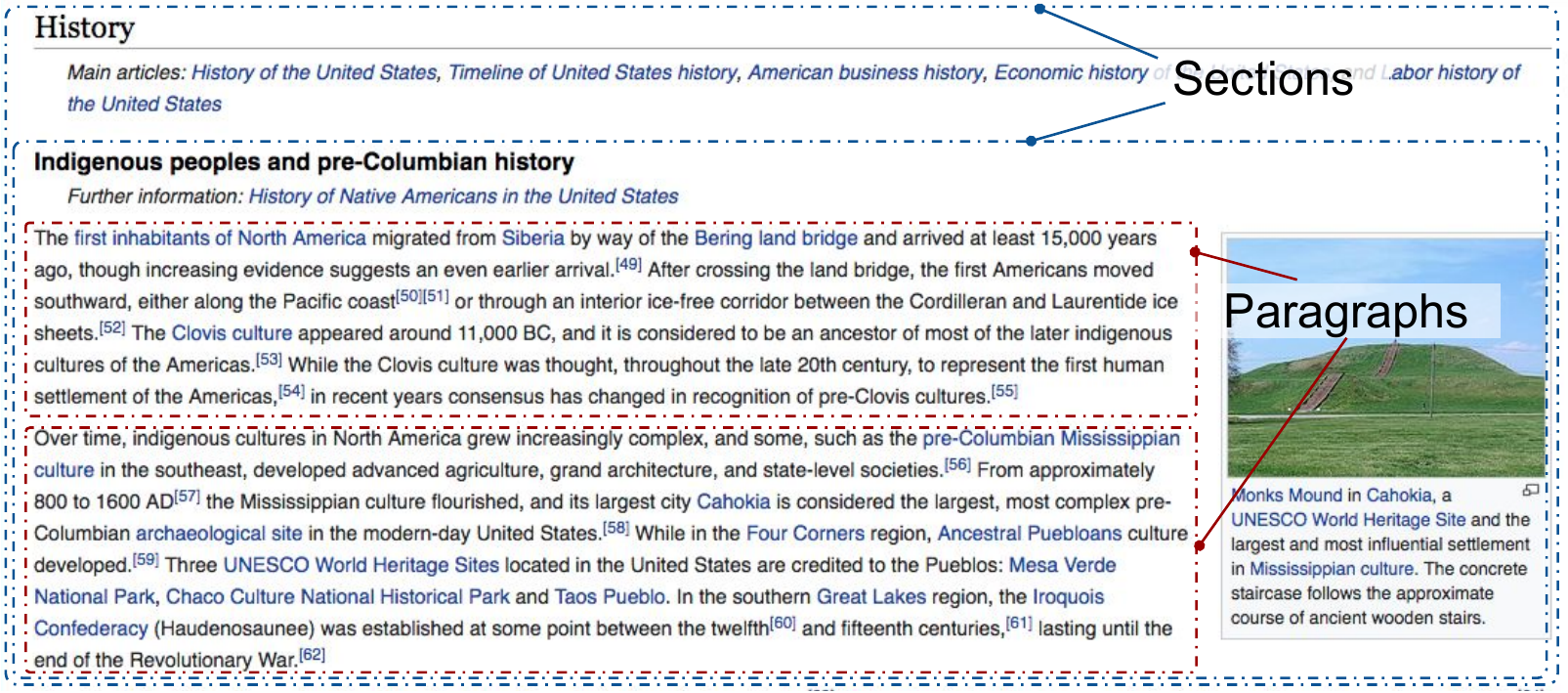}
\caption{Data model illustrated on the Wikipedia article for United States. Source: \url{https://en.wikipedia.org/wiki/United_States}.}
\label{fig.html-example}
\vspace{-10pt}
\end{figure*}

Although Wikipedia article text can be harvested using the official Wikipedia API\footnote{\url{https://en.wikipedia.org/w/api.php}}, it is not recommended to be used at a large scale due to crawling ethics.
The content behind Wikipedia is also provided as XML dumps\footnote{\url{http://dumps.wikimedia.org/}} where the content is represented using Wikitext\footnote{\url{https://en.wikipedia.org/wiki/Help:Wikitext}} (also known as Wiki markup).
Wikitext is a special markup which defines syntax and keywords to be used by the MediaWiki software to format Wikipedia pages.
Apart from the text formatting, it also provides support for LUA scripts\footnote{\url{https://en.wikipedia.org/wiki/Wikipedia:Lua}} and special Wikipedia templates which further manipulate with the content and prepares it for rendering.
These add-ons add additional complexity to the process of rendering and parsing Wikipedia articles.
Although several tools for rendering Wikipedia articles have been developed, there is no tool that implements all templates and LUA scripts.
On the other hand, the Mediawiki\footnote{\url{http://www.mediawiki.org/}} is the only available tool which produces high-quality content from Wikitext markup, thus, in our work we rely on Mediawiki.
We use Mediawiki to expand all templates in the Wikipedia article and render the HTML for the article.
Next, we clean the HTML page based on pre-defined CSS selectors.
We define three types of selectors:
\begin{itemize}
\item \texttt{search selectors} - to find special HTML elements,
\item \texttt{remove selectors} - to remove specific HTML elements, and
\item \texttt{replace selectors} - to replace specific HTML elements, for example with a newline character.
\end{itemize} 
The CSS selector specification is a manual task and their specification is outsourced to the community.
While there are CSS selectors which are valid for all Wikipedias, there is need to specify specific CSS selectors for individual Wikipedia languages.
Upon successful cleansing of the HTML, also with the help of the search selectors the HTML is traversed and its structure is captured.
Via traversing, the article is split into (sub-)\texttt{sections} and \texttt{paragraphs}, and the text of the article is accumulated.
Figure~\ref{fig.html-example} provides an example.
Note that order of the sections and paragraphs is kept as it is in the corresponding Wikipedia article.
Section \texttt{titles} are also captured and stored.
The paragraphs are further parsed and for every \texttt{<a>} HTML element its start and end offsets, surface form and the URL are captured.
If any table or equation is spotted, then it is also extracted, transformed and stored into MathML\footnote{\url{https://www.w3.org/Math/}}.
Finally, the cleaned textual content for the sections, paragraphs and titles is semantically modeled using NIF \cite{hellmann2013NIF} as described in the following section.

\subsection{Data Model in NIF}
\label{sec.nif-model}

The NLP Interchange Format (NIF)\footnote{\url{http://persistence.uni-leipzig.org/nlp2rdf/}}\cite{hellmann2013NIF} is an RDF/OWL-based format that aims to achieve interoperability between Natural Language Processing (NLP) tools, language resources and annotations.
It enables modeling of text documents and describing strings within the documents.
In our work we use selected subset of the NIF concepts to model the extracted content and its structure.
The \textbf{document} and the main content are represented using the \texttt{nif:Context}\footnote{In this article, the \texttt{nif} prefix is associated with the URI \url{http://persistence.uni-leipzig.org/nlp2rdf/ontologies/nif-core\#}.}, while the associated \texttt{nif:String} property holds the actual text.
The \texttt{nif:beginIndex} and \texttt{nif:endIndex} properties are used to describe the position of the sting using offsets and denote its length.
The provenance information is captured with \texttt{nif:sourceUrl}  where the URL identifies the document the context was extracted from and the \texttt{nif:predLang} describes the predominant language of the text.

The \textbf{structure} of the document is captured with the \texttt{nif:Section} and \texttt{nif:Paragraph} classes.
From the \texttt{nif:Context} is referenced the first section using \texttt{nif:firstSection}, the last section with \texttt{nif:lastSection}, and all other contained sections with the \texttt{nif:hasSection} property.
In a same manner subsection within a section can be modeled.
Sections have also begin and end index and they are referenced to the \texttt{nif:Context} using the \texttt{nif:referenceContext} property.
Each section provides references to all contained paragraphs using the \texttt{nif:firstParagraph}, \texttt{nif:lastParagraph} and \texttt{nif:hasParagraph} property.
Paragraphs, same as the sections, are described with a begin and end index and are referenced to the context with the \texttt{nif:referenceContext} property.
Reference to the section is provided with the \texttt{nif:superString} property, which is used to express that one string is contained in another.

The \textbf{links} extracted from the wiki text are described using the \texttt{nif:Word} and \texttt{nif:Phrase} classes.
The \texttt{nif:Word} describes a link with a single-token anchor text, while the \texttt{nif:Phrase} describes a link with a multi-token anchor text.
The anchor text for the link is provided with the \texttt{nif:anchorOf} together with its position in the string of the referenced \texttt{nif:Context}.
Each link resource provides reference to the \texttt{nif:Context} using \texttt{nif:referenceContext} and to the \texttt{nif:Paragraph} using the \texttt{nif:superString}.
The actual link (URL) is described with the \texttt{itsrdf:taIdentRef} property from the Internationalization Tag Set (ITS) Version 2.0\footnote{\url{https://www.w3.org/TR/its20/}} standard.
Following listing provides an RDF excerpt from the datasets which represents the content and the structure of the HTML from Figure~\ref{fig.html-example}.
It describes a section, the first paragraph withing the section and a link.

\begin{lstlisting}[captionpos=b,
language=ttl,
showstringspaces=false,
stepnumber=1,
basicstyle=\scriptsize\ttfamily,
xleftmargin=\parindent,
numbers=left,
breaklines=true,
numberstyle=\scriptsize,
caption=Excerpt from the dataset illustrating the RDF generated for the example from Figure~\ref{fig.html-example}.,
keywordstyle=\color{lightblue}]
@prefix rdf:   <http://www.w3.org/1999/02/22-rdf-syntax-ns#> .
@prefix xsd:   <http://www.w3.org/2001/XMLSchema#> .
@prefix itsrdf: <http://www.w3.org/2005/11/its/rdf#> .
@prefix nif: <> .
@prefix ex: <http://nif.dbpedia.org/wiki/en/> .

ex:United_States?dbpv=2016-10&nif=context
  a	nif:Context ;
  nif:beginIndex	"0"^^xsd:nonNegativeInteger ;
  nif:endIndex		"104211"^^xsd:nonNegativeInteger ;
  nif:firstSection	ex:United_States?dbpv=2016-10&char=0,4241 ;
  nif:lastSection	ex:United_States?dbpv=2016-10&char=103211,104211 ;
  nif:hasSection	ex:World_War_II?dbpv=2016-10&char=0,5001 ;
  nif:sourceUrl		ex:United_States?oldid=745182619 ;
  nif:predLang		<http://lexvo.org/id/iso639-3/eng> ;
  nif:isString	"...The first inhabitants of North America migrated from Siberia by way of the Bering land bridge ..." .

ex:United_States?dbpv=2016-10&char=7745,9418
  a	nif:Section ;
  nif:beginIndex	"7745"^^xsd:nonNegativeInteger ;
  nif:endIndex		"9418"^^xsd:nonNegativeInteger ;
  nif:hasParagraph	ex:United_States?dbpv=2016-10&char=7860,8740 ;
  nif:lastParagraph	ex:United_States?dbpv=2016-10&char=8741,9418 ;
  nif:nextSection	ex:United_States?dbpv=2016-10&char=9420,12898 ;
  nif:referenceContext	ex:United_States?dbpv=2016-10&nif=context ;
  nif:superString	ex:United_States?dbpv=2016-10&char=7548,7743 .

ex:United_States?dbpv=2016-10&nif=paragraph&char=7860,8740
  a	nif:Paragraph ;
  nif:beginIndex	"7860"^^xsd:nonNegativeInteger ;
  nif:endIndex		"8740"^^xsd:nonNegativeInteger ;
  nif:nextParagraph	ex:United_States?dbpv=2016-10&char=8741,9418 ;
  nif:referenceContext	ex:United_States?dbpv=2016-10&nif=context ;
  nif:superString	ex:United_States?dbpv=2016-10&char=7745,9418 .
        
ex:United_States?dbpv=2016-10&char=7913,7920
  a	nif:Word ;
  nif:anchorOf		"Siberia" ;
  nif:beginIndex	"7913"^^xsd:nonNegativeInteger ;
  nif:endIndex		"7920"^^xsd:nonNegativeInteger ;
  nif:referenceContext	ex:United_States?dbpv=2016-10&nif=context ;
  nif:superString	ex:United_States?dbpv=2016-10&char=7860,8740 ;
  itsrdf:taIdentRef	<http://dbpedia.org/resource/Siberia> .
                
\end{lstlisting}
\label{lst.nif-example}

\subsection{Coverage and Availability}
The DBpedia NIF dataset is first of its kind with structured information provided for 128 Wikipedia languages.
The DBpedia NIF dataset provides over 9 billion triples, which is bringing the overall DBpedia triples count up to 23 billion.
Table~\ref{tbl.general-dataset-info} provide statistics for the top 10 Wikipedia languages.
It provides information on the number of articles, paragraphs, annotations, and mean and median number of annotations per article.

\begin{table}[!h]
	\caption{Dataset statistics for the top 10 languages.}
	\begin{center}
\begin{tabular}{ 
>{\raggedleft\arraybackslash} m{17mm} 
>{\raggedleft\arraybackslash} m{17mm}
>{\raggedleft\arraybackslash} m{20mm}
>{\raggedleft\arraybackslash} m{22mm}
>{\centering\arraybackslash} m{18mm}
>{\centering\arraybackslash} m{18mm}
}
\specialrule{1pt}{0pt}{1pt}
\textbf{Language} & 
\textbf{Articles}  & 
\textbf{Paragraphs} &
\textbf{Links} &
\textbf{Mean per article} &
\textbf{Median per article} \\

\specialrule{1pt}{1pt}{3pt}
English & 4,909,454 & 40,939,057 & 127,227,173 & 26.20 & 13 \\
Cebuano & 3,071,209 & 6,434,998 & 24,878,067 & 8.14 & 7 \\
Swedish & 3,297,038 & 7,918,709 & 36,372,087 & 11.07 & 9 \\
German & 1,734,835 & 14897948 & 50,116,852 & 29.08 & 19 \\
French & 1,680,645 & 15,833,816 & 55,347,176 & 33.73 & 17 \\
Dutch & 1,799,619 & 6,537,238 & 23,107,130 & 13.17 & 4 \\
Russian & 1,172,548 & 10,327,544 & 31,759,092 & 27.37 & 15 \\
Italian & 1,124,751 & 7,837,457 & 30,996,231 & 27.79 & 12 \\
Spanish & 1,166,614 & 9,221,266 & 31,123,375 & 26.89 & 16 \\
Polish & 1,106,247 & 5,740,870 & 21,793,337 & 19.84 & 12 \\

\specialrule{1pt}{1pt}{1pt}

Total & 21062960 & 125688903 & 432720520 & 223.28 & 124 \\

\specialrule{1pt}{0pt}{1pt}
\end{tabular}
	\end{center}
	\label{tbl.orig-dataset-stats}
\end{table}

According to the statistics presented in Table~\ref{tbl.general-dataset-info}, although the English Wikipedia is the largest, the German and French Wikipedia articles are enriched with more links in average compared to the English Wikipedia.
French Wikipedia contains 33.73 links per article (mean count) , German Wikipedia 29.08, while English Wikipedia 26.20.
Also, although the Cebuano Wikipedia is the second largest Wikipedia according to the number of articles, it is the one with the least number of links, likely due to the nature of its creation; automated creation of the articles using bot\footnote{\url{https://www.quora.com/Why-are-there-so-many-articles-in-the-Cebuano\%2Dlanguage-on-Wikipedia}}.

Since the amount of data is considerable large, only the English subset of the dataset is published according to the Linked Data principles with dereferenceable URIs.
Publishing of other language versions of the dataset will be consider only if there is requirement for this within the community.
For all resources we mint URIs in the DBpedia namespace (http://nif.dbpedia.org/wiki/\{lang\}/\{name\}).
This namespace gives us flexibility in publishing different language versions of the dataset, as well as, publishing any other NIF dataset.

Information about the latest news, releases, changes and download information is provided at the main dataset page at \url{http://wiki.dbpedia.org/nif-abstract-datasets}.
General information about the dataset is provided in Table~\ref{tbl.general-dataset-info}.

\begin{table}[!h]
	\caption{Details for the DBpedia NIF dataset.}
	\begin{center}
\begin{tabular}{ 
>{\raggedright\arraybackslash} m{27mm} 
>{\raggedright\arraybackslash} m{87mm} 
}
\specialrule{1pt}{0pt}{1pt}
\textbf{Name} & DBpedia NIF dataset\\
\textbf{URL}  & \url{http://wiki.dbpedia.org/nif-abstract-datasets}\\
\textbf{Ontology} & NLP Interchange Format (NIF) version 2.1\\
\textbf{Version} & 1.0\\
\textbf{Release Date} & July 2017\\
\textbf{License} &  Attribution-ShareAlike 4.0 International (CC BY-SA 4.0) \\

\specialrule{1pt}{0pt}{1pt}
\end{tabular}
	\end{center}
	\label{tbl.general-dataset-info}
\end{table}

Currently, the DBpedia NIF dataset is released together with the regular bi-annual DBpedia releases.
Nevertheless, in the past few months the core DBpedia team has done considerable amount of work to streamline the extraction process of DBpedia and convert many of the extraction tasks into an ETL setting.
The ultimate goal behind these efforts is to increase the frequency of the DBpedia releases including the DBpedia NIF dataset.

\subsection{Maintenance and Sustainability Plans}
The DBpedia Association provided us with computational resources for creation of the dataset and a persistent web space for hosting.
This will guarantee persistent URI identifiers for the dataset resources.
The ongoing maintenance of the dataset is an iterative process and feedback from its users is captured via several communication channels: the DBpedia-discussion mailing list\footnote{\url{https://lists.sourceforge.net/lists/listinfo/dbpedia-discussion}}, the TextExt challenge (see next section) and the DBpedia Framework issue tracker\footnote{\url{https://github.com/dbpedia/extraction-framework/}}.

In order to assure sustainable and regularly updated and maintained dataset, the \textbf{TextExt: DBpedia Open Extraction Challenge}\footnote{\url{http://wiki.dbpedia.org/textext}} has been developed.
\textit{TextExt} is a knowledge extraction challenge with the ultimate goal to spur knowledge extraction from Wikipedia article texts in order to dramatically broaden and deepen the amount of structured DBpedia/Wikipedia data and provide a platform for training and benchmarking various NLP tools.
It is a continuous challenge with the focus to sustainably advance the state of the art and systematically put advance knowledge extraction technologies in action.
The challenge is using nine language versions of the DBpedia NIF dataset and participants are asked to apply their knowledge extraction tools in order to extract i) facts, relations, events, terminology or ontologies for the \textit{triples track}, ii) and NLP annotations such as pos-tags, dependencies or co-references for the \textit{annotations track}.
The submissions are evaluated two times a year and the challenge committee selects a winner which receives a reward.
The knowledge extraction tools developed by the challenge participants are executed in regular intervals and the extracted knowledge published as part of the DBpedia core dataset.

\section{Dataset Enrichment}
\label{sec.enrichment}

Wikipedia defines guidelines which should be followed by the authors when authoring content and adding links.
The main principles described in the guidelines\footnote{\url{https://en.wikipedia.org/wiki/Wikipedia:Manual_of_Style/Linking\#Principles}} are as follows:
\begin{enumerate}
\item a link should appear only once in an article,
\item avoid linking subjects known by most readers,
\item avoid overlinking which makes it difficult to identify useful links, and
\item add links specific to the topic of the article.
\end{enumerate}

These guidelines manifest themselves in certain properties of the dataset:
\begin{enumerate}
\item If a concept is linked once, its further occurrences will not be linked.
\item The concept described in the article is never linked to itself.
\item Relevant subjects related to the article are to be linked.
\end{enumerate}

While the third property is important for the dataset, the first and the second properties negatively influence the number of links and the richness of the dataset.
In our work, we addressed these two properties and further enriched the dataset with links.
The main goal of the enrichment process is to annotate those words or phrases which have been annotated at least once in the article but their subsequent occurrences lack links.

The enrichment workflow which creates new links within the Wikipedia articles has been defined as follows.
First, all links and their corresponding anchor texts found in the article are collected.
Next, the link-anchor pairs are sorted according to the anchor text length, from the longest to the shortest anchor.
Following the ordered list, a full string matching is applied over the whole article content, starting with a lookup of anchors with highest text length.
In case of an overlapping matches, the one with the longest match will have the priority and the other is discarded.
For example, if a link with anchor text ``East Berlin'' already exists, then a link over ``Berlin'' will be omitted.

We have applied the enrichment process on the content extracted from the top ten Wikipedias, ranked according to the number of articles.
Note that enrichment within the ``See also'', ``Notes'', ``Bibliography'', ``References'' and ``External Links'' section have not been applied.
Table~\ref{tbl.enrichment-stats} summarizes the results from the enrichment.

\begin{table}[!t]
	\caption{Dataset enrichment statistics.}
	\begin{center}
\begin{tabular}{ 
>{\raggedleft\arraybackslash} m{16mm} 
>{\raggedleft\arraybackslash} m{25mm}
>{\raggedleft\arraybackslash} m{25mm}
>{\raggedleft\arraybackslash} m{25mm}
>{\raggedleft\arraybackslash} m{25mm}
}
\specialrule{1pt}{0pt}{1pt}
\textbf{Language} & 
\multicolumn{1}{>{\centering\arraybackslash}m{23mm}}{\textbf{Annotations before enrichment}}  & 
\multicolumn{1}{>{\centering\arraybackslash}m{23mm}}{\textbf{Unique annotations}} &
\multicolumn{1}{>{\centering\arraybackslash}m{23mm}}{\textbf{Annotations after enrichment}}  & 
\multicolumn{1}{>{\centering\arraybackslash}m{23mm}}{\textbf{\% of new annotations}} \\

\specialrule{1pt}{1pt}{0pt}

English & 127,227,173 & 17,322,066 & 168,988,631 & 31.36 \%  \\
Cebuano & 24,878,067 & 3,077,130 & 26,222,416 & 5.40 \% \\
Swedish & 36,372,087 & 2,974,858 & 41,368,833 & 13.74 \% \\
German & 50,116,852 & 6,642,511 & 63,347,163 & 26.39 \%\\
French & 55,347,176 & 6,110,952 & 74,843,900 & 35.23 \%\\
Dutch & 23,107,130 & 3,332,920 & 26,873,294 & 16.29 \%\\
Russian & 31,759,092 & 5,386,638 & 37,215,365 & 17.18 \%\\
Italian & 30,996,231 & 3,674,701 & 39,605,915 & 27.78 \%\\
Spanish & 31,123,375 & 4,116,251 & 40,125,126 & 28.92 \%\\
Polish & 21,793,337 & 3,532,420 & 24,374,548 & 11.84 \%\\

\specialrule{1pt}{1pt}{1pt}

Total & 432,720,520 & 56,170,447 & 542,965,191 & 25.48 \%\\

\specialrule{1pt}{0pt}{1pt}
\end{tabular}
	\end{center}
	\label{tbl.enrichment-stats}
\end{table}

In overall, the enrichment process generated over 110 million links which is more than 25\% more links compared to the number of links present in Wikipedia.
Most new links have been generated for the French Wikipedia 35.23\%, followed by the English, 31.36\%, Spanish 28.92\%, Italian 27.78\% and German 26.39\%.
The reason for the different percentage of generated new links could be the language, how strong the guidelines have been followed by the Wikipedia editors, or the cultural background of the editors.
For example, the content of the Cebuano Wikipedia has been primarily generated automatically using bot, which has direct influence on the enrichment process (only 5.4\% new links.) due to the low number of initial links present in the content.
We have evaluated the enrichment process and report on the results from the evaluation in Section~\ref{sec.sem-quality-eval}.

\section{Dataset Quality}
\label{sec.quality}

According to the 5-star dataset classification system defined by Tim Berners-Lee \cite{berners2006linked}, the DBpedia NIF dataset classifies as a five-star dataset.
The five stars are credited for the open license, availability in a machine-readable format, use of open standards, use of URIs for identification, and the links to the other LOD datasets; links to DBpedia and indirectly to many other datasets.

In \cite{zaveri2015quality} the authors describe a list of indicators for evaluation of the intrinsic quality of Linked Data datasets.
We have checked the data for each of the metrics (where applicable) described for syntactic validity and semantic accuracy.

\subsection{Syntactic Validity of the Dataset}
\label{sec.syn-quality-eval}
Considering the size and the complexity of the task in extraction knowledge from HTML, we have put an significant effort in checking the syntactic validity of the dataset.
Following three syntactic validation checks have been executed:

\begin{itemize}

\item \textit{Raptor}\footnote{\url{http://librdf.org/raptor/}} RDF syntax parsing and serializing utility was used to assure the RDF is syntactically correct.

\item The GNU command line tools \textit{iconv}\footnote{\url{https://www.gnu.org/software/libiconv/}} and \textit{wc}\footnote{\url{https://www.gnu.org/software/coreutils/manual/html_node/wc-invocation.html}} have been used to make sure the files only contain valid unicode codepoints.
Wrong encoded characters were dropped and we also compared the number of characters afterwards to the character counts of the original files.
Iconv was executed on the files with the \texttt{-c} parameter which drops inconvertible and wrongly encoded characters.

\item \textit{RDFUnit}\footnote{\url{https://github.com/AKSW/RDFUnit/}}\cite{Kontokostas2014} is a validation tool designed to read and produce RDF that complies to an ontology.
RDFUnit  has been used to ensure adherence to the NIF format \cite{hellmann2013NIF}.
Tests, such as checking that the string indicated with the \texttt{nif:anchorOf} property equals the part of \texttt{nif:isString} string (article text) found between begin and end offset, have been applied.

\end{itemize}

\subsection{Semantic Accuracy of the Dataset Enrichment}
\label{sec.sem-quality-eval}
In order to evaluate the semantic accuracy of the enrichment process, we have crowdsourced randomly created subset of the dataset and asked annotators to check the enrichments; the newly generated links.
The goal of this evaluation was to check the quality of the enrichments, and in particular to assess i) if the anchors (annotations) are at correct position within the text, and ii) whether the link associated with the anchor text is correct.

\noindent \textbf{Evaluation setup.} For the evaluation we have used the top-10 articles for English and German according to their PageRank score.
\textit{Notice: for the review purposes we have temporarily published these documents (original and enriched version) at} \url{http://nlp2rdf.aksw.org/dbpedia-nif-examples/}.
The score has been retrieved from the DBpedia SPARQL endpoint\footnote{\url{http://dbpedia.org/sparql}}, found in the \texttt{\url{http://people.aifb.kit.edu/ath/\#DBpedia_PageRank}} graph and described with the \texttt{\url{http://purl.org/voc/vrank\#rankValue}} property.
Since the number of links in these articles is high (over 1,500 links per article), for the evaluation we have decided to randomly select 30 links per article.
Next, these articles were submitted to ten annotators, where each annotator processed at minimum 150 links from five articles.
For each annotation we collected three judgments.
The evaluation sheets contained list of enrichments described with an ``\texttt{anchor text}'', the ``\texttt{link}'' associated with the anchor, and a ``\texttt{context}'' text (15 words) which contains the link.
Since the provided context text is short in length, we also provided an HTML version of the article with the links highlighted in it.
Along with the evaluation sheets we have also provided strict annotation guidelines where each annotator was asked to check if:
\begin{itemize}
\item \textbf{anchor-eval (0 or 1):} the anchor delimits the complete word or phrase that identifies an entity/concept/term in relation to the context,
\item \textbf{link-eval (0 or 1):} the link (i.e. Wikipedia page) describes the anchor,
\item \textbf{comment:} a placeholder for an optional comment.
\end{itemize}

The annotators were instructed to provide explanation in the \textit{comment} field, if they felt unsure.

The composition of annotators was as follows: five computer science PhD students, four undergraduate bachelor computer science students and one master of arts student with a strong knowledge in computer science.
All annotators were non-native English speakers and eight of them native German speakers.

\textbf{Results.} Table~\ref{tbl.enrichment-eval-results} shows the results from the evaluation.
We report on the fraction of correctly generated new links among the total number of generated links; a link is considered to be marked as correct if at least two out of three annotators provided same judgment.
We also report on the correctly identified anchors, links, and their combination. 
In addition, for each evaluation mode we report the inter-annotator agreement (IAA) in terms of Fleiss' kappa \cite{fleiss1971measuring} metric.

\newcolumntype{P}[1]{>{\centering\arraybackslash}p{#1}}

\begin{table}[!h]
	\caption{Results from the evaluation of the dataset enrichment.}
	\begin{center}
\begin{tabular}{ 
P{1.1cm}
P{1.5cm}
P{1.5cm}
P{1.5cm}
P{1.5cm}
P{1.5cm}
P{1.5cm}
}
\specialrule{1pt}{0pt}{1pt}
& 
\multicolumn{1}{>{\centering\arraybackslash}m{15mm}}{\textbf{IAA for anchors}} &
\multicolumn{1}{>{\centering\arraybackslash}m{15mm}}{\textbf{IAA for links}} &
\multicolumn{1}{>{\centering\arraybackslash}m{20mm}}{\textbf{IAA for anchors and links}} &
\multicolumn{1}{>{\centering\arraybackslash}m{15mm}}{\textbf{Correct anchors}}  & 
\multicolumn{1}{>{\centering\arraybackslash}m{15mm}}{\textbf{Correct links}}  & 
\multicolumn{1}{>{\centering\arraybackslash}m{20mm}}{\textbf{Correct anchors and links}} \\

\specialrule{1pt}{1pt}{3pt}

English & 0.4929 &  0.6538 & 0.5435 & 0.7933 & 0.6633  & 0.6133 \\

\specialrule{1pt}{1pt}{3pt}

German & 0.4828 & 0.6183 & 0.5177 & 0.8723 & 0.8511 & 0.7908 \\

\specialrule{1pt}{1pt}{1pt}

\end{tabular}
	\end{center}
	\label{tbl.enrichment-eval-results}
\end{table}

According to the agreement interpretation table defined in \cite{landis1977measurement}, we have received ``substantial agreement'' (0.61-0.80) for the links; 0.6538 for English and 0.6183 for German.
For the rest, the anchors and both, the anchors and links combined, we have received ``moderate agreement'' (0.41-0.60).
Highest agreement score has been received for the links in English 0.6538, while lowest for the German anchors 0.4828.

The results also show that the enrichment process performed well in terms of accuracy.
Best results have been achieved for German anchors with fraction of 0.8723 correct anchors, while the worst for the English links with fraction of 0.6633 correct links.
We assume that the low fraction of correct links for English is due to the character of the Wikipedia guidelines on linking.
In our future work, we will focus on improvement of the enrichment process, with a particular focus on the link validation.

\section{Selected Dataset Use Cases from the TextExt Challenge}
\label{sec.use-cases}
In this section we report on two use cases of the datasets which won the TextExt challenge in year 2017.

\subsection{Linked Hypernyms Dataset: entity typing using Wikipedia}
The Linked Hypernyms Dataset (LHD) \cite{KLIEGR201647} was the first winner at the TextExt challenge which has been using the DBpedia NIF dataset.
The LHD dataset has been developed on top of the DBpedia NIF dataset with the goal to complete missing types in DBpedia and mine more specific types.
It has been generated via extraction of the hypernyms found in the first sentence in the articles.
The extracted hypernyms are considered as entity type for the entity described in the article.
Finally, the hypernyms are mapped to DBpedia ontology types which are further analyzed and the most specific type selected.
LHD has been processed for English, German and Dutch and it is already integrated as part of the DBpedia core dataset.

\subsection{Lector: facts extraction from Wikipedia text}
Lector \cite{Cannaviccio:2016} was the second and most recent winner of the TextExt challenge.
Lector is a knowledge extraction tool used to harvest highly accurate facts from Wikipedia article text.
The tool has been adopted and applied on DBpedia NIF to extract new facts.
The approach applied on DBpedia NIF is defined as follows: in the first phase, all articles are processed in order to harvest the most common typed-phrases used to express facts present in DBpedia; in the second phase, the most common phrases are used to extract novel facts.
Lector has been applied on five different DBpedia NIF languages and in a near future it will be integrated as part of the DBpedia core dataset.

\section{Conclusions}
\label{sec.conclusion}
Unstructured Wikipedia article texts provide vast amount of valuable information, which has not yet been exploited and considered in DBpedia.
In this paper, we have presented the \textit{DBpedia NIF} dataset, a large scale, open and multilingual knowledge extraction corpus.
The ultimate goal of the dataset is to broaden and deepen the information in DBpedia and to provide large-scale and multilingual linguistic resource for training various NLP and IR tasks.
The dataset provides the content of Wikipedia article texts in 128 languages and it captures the structure of the articles as they are structured in Wikipedia including the sections, paragraphs and links.
We use the NIF format to model the data and represent the information.
Also, we have further enriched the dataset and increased the number of links in the content by 25\%.
The dataset has been checked for its syntactic validity and semantic accuracy and the results from these evaluations confirm the quality of the dataset.
We have also validated and shown the potential of the dataset on two selected use cases from the TextExt challenge.
Presented maintenance and sustainability plans will assure continuous growth of the dataset in terms of size, quality, coverage and usage.
In our future work, we will focus on improving the extraction process and the data model, and continue to disseminate and extract additional knowledge from the dataset at the TextExt challenge.

~

\noindent \textbf{Acknowledgement.} We thank all contributors to the dataset and especially Martin Br\"ummer for the initial implementation and Markus Freudenberg for integration of the extraction as part of the DBpedia Extraction framework.
Also, thanks to the annotators for their help with the evaluation.
This work was partially funded by a grant from the EU’s H2020 Programme for the ALIGNED project (GA-644055) and grant from the Federal Ministry for Economic Affairs and Energy of Germany (BMWi) for the SmartDataWeb project (GA-01MD15010B).

%
%

\bibliographystyle{plainnat}

\end{document}